**Title:** Towards Outcome-Driven Patient Subgroups: A Machine Learning Analysis Across Six Depression Treatment Studies

Authors: David Benrimoh[1,2,3]\*, Akiva Kleinerman[4]\*, Toshi A. Furukawa[5], Charles F. Reynolds III[6,7], Eric Lenze[8], Jordan Karp[9], Benoit Mulsant[10], Caitrin Armstrong[3], Joseph Mehltretter[3], Robert Fratila[3], Kelly Perlman[1,3], Sonia Israel[3], Myriam Tanguay-Sela[12], Christina Popescu[3], Grace Golden[3], Sabrina Qassim[3], Alexandra Anacleto[3], Adam Kapelner[11], Ariel Rosenfeld[4], Gustavo Turecki[1]

\*DB and AK contributed equally to this paper

Affiliations:
1. McGill University, Department of Psychiatry, Montreal, Canada
2. Stanford University, Department of Psychiatry, Stanford, USA
3. Aifred Health, Montreal, Canada
4. Bar-Ilan University, Ramat Gan, Israel
5. Department of Health Promotion and Human Behavior, Kyoto University Graduate School of Medicine / School of Public Health
6. Department of Psychiatry, University of Pittsburgh School of Medicine
7. Department of Psychiatry, Tufts University School of Medicine
8. Department of Psychiatry at Washington University School of Medicine in St. Louis, Missouri
9. Department of Psychiatry, University of Arizona
10. Department of Psychiatry, University of Toronto
11. Department of Mathematics, Queens College, City University of New York
12. Université du Québec à Montréal, Department of Psychology

Corresponding author: David Benrimoh, MD.CM., MSc., MSc., FRCPC
David.benrimoh@mail.mcgill.ca
Stanford Department of Psychiatry, 401 Quarry Road, Stanford, CA 94305
514-464-7813

Word count: 3,600

# Abstract:

**Importance:** Major depressive disorder (MDD) is a heterogeneous condition; multiple underlying neurobiological substrates could be associated with treatment response variability. Understanding the sources of this variability and predicting outcomes has been elusive. Machine learning (ML) has shown promise in predicting treatment response in MDD, but one limitation has been the lack of clinical interpretability of machine learning models, limiting clinician confidence in model results.
**Objective:** To develop a machine learning model to derive treatment-relevant patient profiles using clinical and demographic information.
**Design:** We analyzed data from six clinical trials of pharmacological treatment for depression (total n = 5438) using the Differential Prototypes Neural Network (DPNN), a neural network model that derives patient prototypes which can be used to derive treatment-relevant patient clusters while learning to generate probabilities for differential treatment response. A model classifying remission and outputting individual remission probabilities for five first-line monotherapies and three combination treatments was trained using clinical and demographic data.
**Setting:** Previously-conducted clinical trials of antidepressant medications.
**Participants:** Patients with MDD.
**Main outcomes and measures:** Model validity and clinical utility were measured based on area under the curve (AUC) and expected improvement in sample remission rate with model-guided treatment, respectively. Post-hoc analyses yielded clusters (subgroups) based on patient prototypes learned during training. Prototypes were evaluated for interpretability by assessing differences in feature distributions (e.g. age, sex, symptom severity) and treatment-specific outcomes.
**Results:** A 3-prototype model achieved an AUC of 0.66 and an expected absolute improvement in population remission rate of 6.5% (relative improvement of 15.6%). We identified three treatment-relevant patient clusters. Cluster A patients tended to be younger, to have increased levels of fatigue and more severe symptoms. Cluster B patients tended to be older, female with less severe symptoms, and the highest remission rates. Cluster C patients had more severe symptoms, lower remission rates, more psychomotor agitation, more intense suicidal ideation, more somatic genital symptoms, and showed improved remission with venlafaxine.
**Conclusion and Relevance:** It is possible to produce novel treatment-relevant patient profiles using machine learning models; doing so may improve precision medicine for depression. **Note: This model is not currently the subject of any active clinical trials and is not intended for clinical use.**

Key points:
**Question:** Can machine learning models of depression treatment response be trained to generate treatment-relevant subgroups from a pool of patients with major depression?
**Findings:** Using the Differential Prototypes Neural Network to analyze six studies of antidepressant medication treatment (n = 5438), we trained a model with the potential to improve population remission rates by 6.5% (15.6% relative improvement). The model generated three novel patient subgroups. These subgroups differed from each other in terms of symptoms (such as psychomotor agitation) and demographic characteristics.
**Meaning:** Machine learning models can be trained to generate treatment-relevant patient subgroups, thereby potentially improving the ability to personalize treatment for depression.

# 1. Introduction:

Major depressive disorder (MDD) affects more than one in nine people (Bromet et al., 2011) and can be challenging to treat as only one in three patients will achieve remission following their first treatment, and one third will still not reach remission following their fourth (Warden et al. 2007). One potential cause of the low treatment success rate is the heterogeneity of depression: two patients with MDD can have very different symptom profiles and respond differently to treatment, presumably due in part to different underlying neurobiology (Benrimoh et al., 2018). Variable etiopathogenesis renders personalizing treatments and the development of new treatments significantly more challenging (Perlman et al., 2019).

There is a lack of readily available tools which can input patients' clinical and demographic data and generate personalized predictions about treatment success, leading many researchers to turn to machine learning to generate predictive models aimed at solving this problem (Squarcina et al., 2021; Chekroud et al., 2016; Iniesta et al., 2016; Mehltretter et al., 2020). In previous work, members of our team demonstrated that, using deep learning, we could train a *differential treatment benefit prediction* model — a model that could compare an arbitrary number of treatments and generate a list of remission probabilities for each treatment for a given patient, with the objective of personalizing treatments even when treatments are roughly equivalent at the population level (Mehltretter et al., 2020).

In addition to generating probabilities of remission for multiple treatments, our previous model also generated an "interpretability report" in order to help clinicians better understand its outputs. This report specified the top five features with the greatest weight for each outputted probability (Mehltretter et al., 2020), analogous to a list of risk and protective factors. This model has been deployed in the form of a clinical decision support system (Benrimoh et al., 2021; Tanguay-Sela et al., 2022; Popescu et al., 2021), and an updated version of the model is currently the subject of an ongoing randomized clinical trial (see **ClinicalTrials.gov Identifier: NCT04655924**). One key observation made during simulation center testing of this system was that the degree to which clinicians used the information provided by the model to make their treatment selection depended on their reported trust in the model. Trust in the model in turn depended on how much clinicians felt that the interpretability report represented the standardized patient with whom they were interacting (Benrimoh et al., 2021; Tanguay-Sela et al., 2022). As such, the interpretability of machine learning models is clearly an important consideration in their clinical utility. The interpretability of a model in a clinical setting may also be relevant to its utility for enhancing shared decision making.

In this study, we sought to demonstrate a novel approach to the interpretability of machine learning outputs for antidepressant treatment choice using our novel neural network architecture, the Differential Prototypes Neural Network (DPNN, Kleinerman et al., 2021). This architecture is designed to learn patient prototypes from which we can derive treatment-relevant subtypes while training to output remission probabilities, and to maximize the difference between these prototypes with respect to the treatment with the greatest observed effectiveness. A prototype can be thought of as a 'typical patient' in the context of a given subgroup. These prototypes, which exist in the model's latent space, can then be translated into descriptive patient clusters (composed of patients who most closely resemble the 'typical patient' for that cluster). Patient clusters represent putative patient subgroups defined on the basis of treatment response, rather than phenomenology alone. In addition to assessing the accuracy of our model and its ability to improve population remission rates, we sought to demonstrate that the DPNN can produce clinically meaningful novel subgroups which can contribute to the interpretability of model outputs. We note that the goal of this study was to address initial treatment choices for a general population of patients with depression.

# 2. Methods

## 2.1. The Datasets

Our data consists of patient-level data from 6 clinical trials of pharmacological treatment for MDD: COMED, STAR*D, REVAMP (focusing on the arms that utilized medications), EMBARC, IRL-GREY, and SUN☺D (Rush et al., 2011; Rush et al., 2006; Trivedi et al., 2008; Trivedi et al., 2016; Lenze et al., 2015; Kato et al., 2018; Yacouby et al., 2020; Loh et al., 2011; Suthaharan et al., 2016). For each dataset we focused on the first treatment utilized; in SUN☺D, where treatments were changed relatively early (at 3 weeks) for those not initially responding to sertraline, we included patients who were randomized to mirtazapine or to mirtazapine-sertraline combination therapy (see supplementary section 1.4).

In order to merge the datasets, it was necessary to identify overlapping features and transform their values so they would have a similar value range. This process is described in detail in the supplementary material (Section 1b). When determining patient remission status, we took an intent-to-treat approach, using all patients and ascertaining binary remission at the last possible measurement (Mehltretter et al., 2020a). Supplementary table 1 presents the treatments and the number of patients who received each treatment.

## 2.2. DPNN - Model Description

The Differential Prototypes Neural Network (DPNN) is a neural network model architecture that aims to generate probabilities of outcomes of multiple treatments for individual patients, while simultaneously distinguishing latent prototypes that vary with respect to the outcomes of specific treatments. It is intended to classify remission while learning to output remission probabilities for different treatments (Mehltretter et al., 2019) and to concurrently generate novel patient prototypes which can serve as seeds for patient subgroups that are defined by their relationship to treatment outcomes. The model is described in detail in (Kleinerman et al., 2021). The model is trained to classify binary remission. The model architecture consists of three main components: 1) an autoencoder that encodes the patient features into a latent space representation; 2) a prototype layer, consisting of a predefined number of patient prototypes, represented in the latent space; and 3) a classification network that is used for obtaining the probability distribution over the possible outcomes for all possible treatments.

2.3. Model Training and Testing

We evaluated 3, 4 and 5 prototype variants of the model. We separated the dataset into a training set and a test set using the k-fold cross validation technique (Rodriguez et al., 2009), in which the dataset is split into k=10 consecutive folds. To generate the split, we used stratification (Zeng & Martinez, 2000) to ensure a similar ratio of the remission to non-remission patients in each fold (i.e. remission rate of close to 41.5%, the sample remission rate). Each fold is then used once for the test set while the k-1 remaining folds are used for the training set. We repeated this process 50 times (shuffling data randomly between folds while maintaining stratification for remission) in order to obtain a sufficiently large results pool to investigate stability of results and subsequently obtained 500 samples of each metric for each prototype number. The final results are an average over the 500 samples, each of which is an average over the k=10 folds. In addition, results from a version of the model holding out one study are presented in supplementary section 6.

2.4. Model evaluation

Our primary evaluation metric was the area under the curve (AUC) for the classification of remission vs. non-remission. We also calculated sensitivity, specificity, and positive and negative predictive values at the model's default decision threshold (50%). We were also interested in determining the degree to which use of model outputted probabilities to guide treatment could increase the remission rate over random allocation (Kleinerman et al., 2021). To accomplish this, we calculated the "remission rate improvement" (RRI) metric. This was the percentage of patients who went into remission among the patients in each test set who actually received the drug with the highest probability of remission as per our model output. This can then be compared to the baseline remission rate.

2.5. Deriving and Interpreting Clusters

The full procedure for generating clusters is presented in figure 1 and supplementary section 4.1. We first trained and tested the DPNN model on the whole dataset. Then we obtained the latent space representation of the prototypes that were tuned by the model during training. We then found, for each prototype, the group of patients that were closest to that prototype in the latent space, based on Euclidean distance. For each cluster, we calculated the average real remission rate for each medication which, in turn, generated a ranking of medications within each cluster based on remission rate. We then visualized the distribution of all feature values in all clusters using a histogram, which allows for comparisons between clusters. We also used these feature distributions to visualize the position of a patient with respect to the derived subgroups, as a visual aid for clinicians. Another visualization method based on decision trees is presented in supplementary section 3.

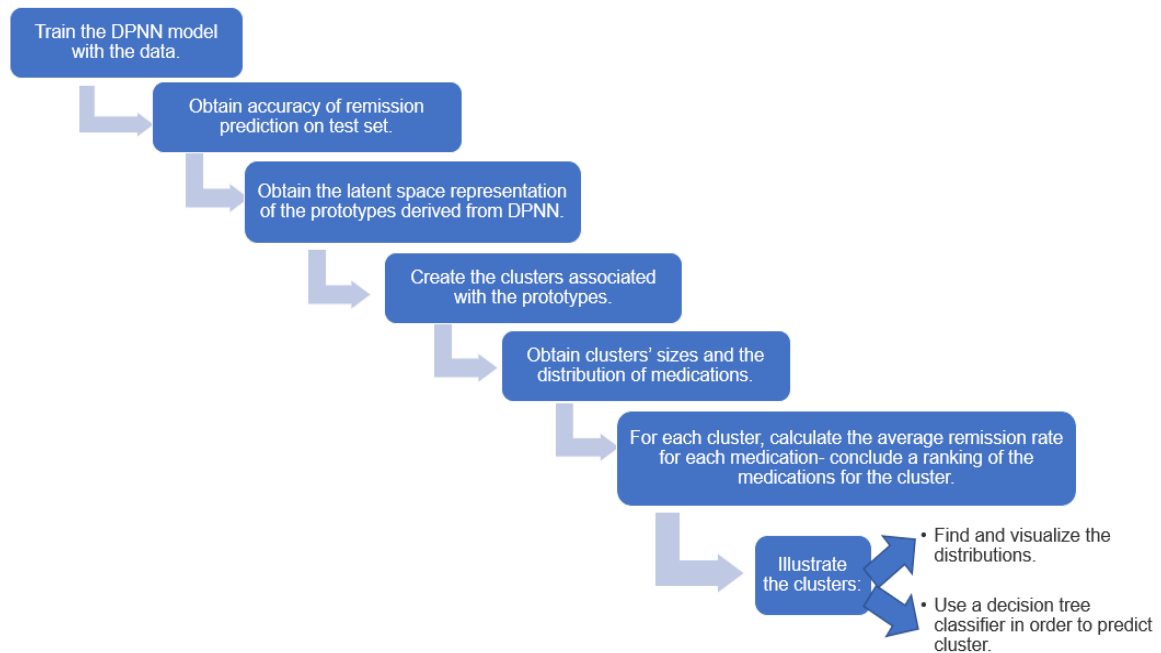

*Figure 1. The process of generating prototypes and clusters from the DPNN model.*

## 3. Results

3.1 Patient Demographics

Our final dataset included 5438 patients. The ages of the patients ranged between 18 and 93, and the mean age was 43.5 (SD = 13.9). 61% were female and 39% were male. 61% of the patients were Caucasian, 22% Asian (mostly from the SUN☺D dataset which was conducted in Japan), 9% were of African descent, and the remaining 8% were from other ethnic origins. As such, this dataset covers a wide age range, and it includes more women than men—typically the case in depression clinical trials and in epidemiological studies (Monahan et al., 2022; Albert, 2015). Our data set also contains a large proportion of non-Caucasian participants, which could enhance our model's generalizability.

3.2 Features

After merging the datasets and including as many features as possible which were reliably present across datasets, we were left with 19 features that described each patient at the start of their assigned treatment. Three features described demographic attributes of the patient: age, sex and race/ethnicity. All the other features described the intensity of various depression symptoms recorded in questionnaires (more details can be found in the supplementary material).

3.3 Model Results

Model performance metrics for classifying remission are presented in Table 1. With respect to AUC, our primary metric, results were similar across the 3-, 4- and 5- prototype models: 3 prototypes = 0.666 (SD=0.07) ; 4 prototypes = 0.667 (SD=0.01); 5 prototypes = 0.671 (SD=0.004). These results are also in line with accuracies seen in other models of MDD treatment response prediction using sociodemographic features (Mehltretter et al., 2019; Chekroud et al., 2016). Based on the RRI measure, the 3-prototype model slightly outperformed the other two. The baseline population remission rate in the whole sample is 41.5%; as such, all of our models are expected to improve remission rates (if patients were prescribed the most effective drug based on outputted model probabilities for them) by at least 5.6% in terms of absolute remission rate (a 13.5% relative improvement) (this is consistent with recent literature (Kessler et al., 2020)), and of 6.5% (a 15.6% relative improvement) when considering the 3 prototype model.

| Number of prototypes | Area under curve (AUC)*** | Sensitivity | Specificity | Positive predictive value (ppv) | Negative predictive value (npv) | F1* | Remission Rate** |
|---|---|---|---|---|---|---|---|
| 3 | 0.666 (0.07) | 0.441 (0.3) | 0.731 (0.18) | 0.385 (0.24) | 0.678 (0.06) | 0.400 (0.26) | 0.480 (0.05) |
| 4 | 0.667 (0.01) | 0.433 (0.28) | 0.747 (0.16) | 0.384 (0.24) | 0.675 (0.05) | 0.406 (0.25) | 0. 477 (0.02) |
| 5 | 0.671 (0.004) | 0.445 (0.26) | 0.57 (0.13) | 0.442 (0.22) | 0.679 (0.05) | 0.430 (0.23) | 0.471 (0.02) |

*Table 1. Model performance metrics for remission classification. The values represent the means for the various accuracy metrics across the 50 runs of the model. The numbers in brackets are the standard deviations. Note that aside from AUC and F1, results are calculated at the model's default decision threshold (50%).*
*\*F1 = the harmonic mean of the recall and precision (Yacouby & Axman, 2020).*

*\*\*Remission rate = the percentage of patients who went into remission among the patients who received the drug with the highest probability for them as output by the model. The reference is the population remission rate of 0.405.*
*\*\*\* AUC range from 0 to 1, the higher the better*

3.4 Interpretability Analysis

Due to the lack of large differences in terms of model metrics between the models with 3, 4 and 5 prototypes, we will focus here on the interpretability analysis of the 3-cluster model.

3.4.1 The Clusters

We derived three clusters: A, B and C. Cluster A included 1742 patients, cluster B included 2459 patients and cluster C included 1237 patients. The remission rates (actual observed remission, not model remission probabilities) within the clusters were 35%, 48% and 34% respectively.

Table 2 presents the real remission rates for each treatment within a cluster, in descending order.

| Cluster A | Cluster B | Cluster C |
|---|---|---|
| Citalopram(0.45) | Escitalopram + Bupropion(0.62) | Venlafaxine(0.49) |
| Venlafaxine + Mirtazapine(0.43) | Venlafaxine(0.59) | Escitalopram(0.39) |
| Escitalopram + Bupropion(0.42) | Citalopram(0.5) | Citalopram(0.36) |
| Venlafaxine(0.4) | Venlafaxine + Mirtazapine(0.49) | Sertraline(0.34) |
| Escitalopram(0.35) | Mirtazapine(0.47) | Escitalopram + Bupropion(0.34) |
| Sertraline(0.34) | Mirtazapine+Sertraline(0.47) | Venlafaxine + Mirtazapine(0.33) |
| Mirtazapine+Sertraline(0.19) | Escitalopram(0.39) | Mirtazapine+Sertraline(0.18) |
| Mirtazapine(0.18) | Sertraline(0.36) | Mirtazapine(0.15) |

*Table 2. Real remission rates for each treatment in each cluster*

### 3.4.2 Feature Distributions

In order to interpret the clusters from a clinical perspective, we analyzed the distribution of the values of all the (non-latent) features and compared these distributions across clusters. We demonstrate key features which differed between clusters in figure 2. The rest of the feature distributions are presented in the supplementary material.

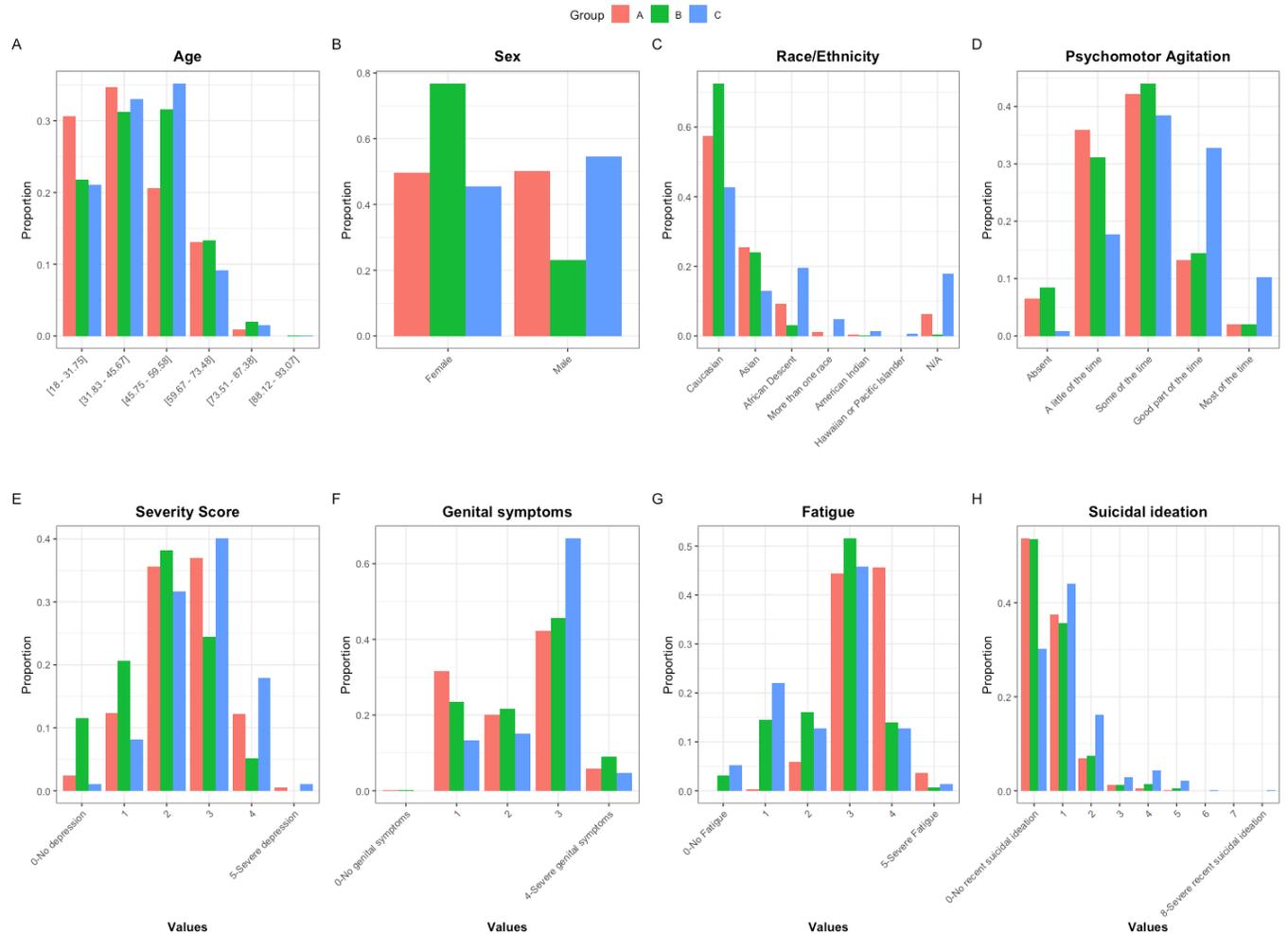

*Figure 2. Distributions for key features, separated by cluster. In most cases, as a result of the re-scaling of features between studies, the feature values were not integers. Therefore, in order to create these charts, we first rounded the original values and then calculated the frequency of the rounded values. Note N/A in race ethnicity denotes missing or not reported.*

Patients in cluster A tend to be younger and to have increased levels of fatigue as well as more severe depression and lower remission rates than cluster B. Patients in cluster B tend to be older and are more likely to be female; this cluster has the highest remission rates, and tends to have less severe symptoms than the other two clusters. Cluster C has lower remission rates than cluster B, as well as more suicidal ideation, more somatic genital symptoms (e.g. loss of libido) and more psychomotor agitation than the other two clusters. Cluster C also tends to be older, and to benefit the most from venlafaxine (49% remission rate), representing the largest jump in remission rate between the first and second treatment in any of the clusters. Cluster C also is the only cluster for which no combination treatments, or treatments including bupropion, appear in the top 3 treatments. In the supplementary material (Subsection 2.2) we present a statistical analysis of the feature differences between the clusters.

In addition to reviewing the data at the group level, in figure 3 we provide a visualization aimed at improving interpretability at the patient level.

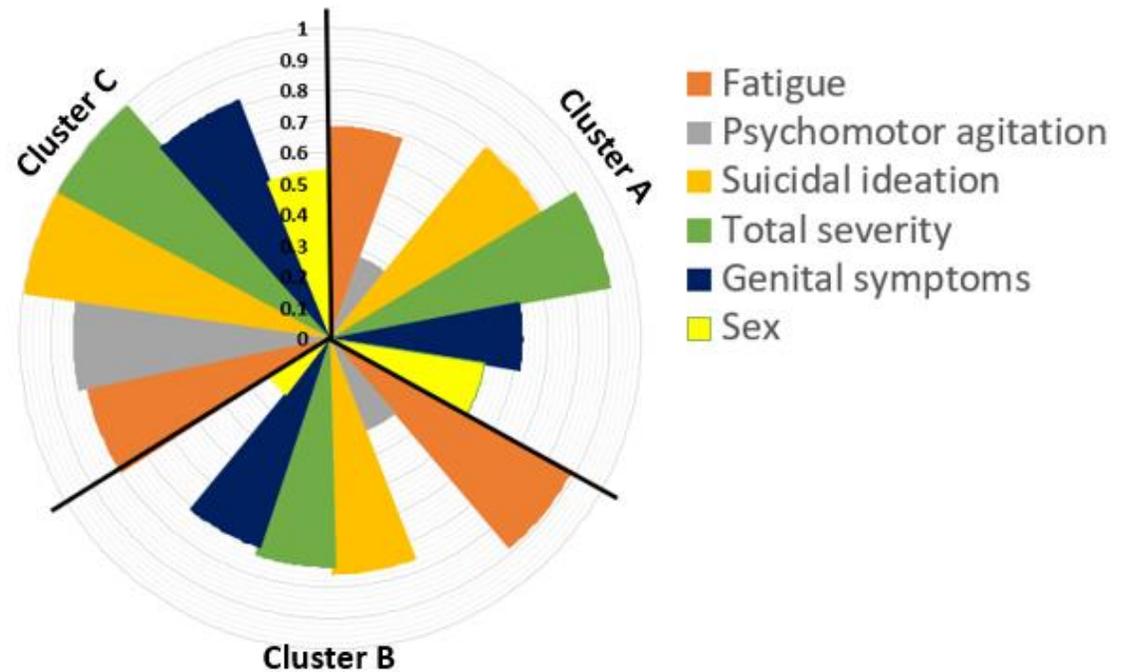

*Figure 3. Visualization of the patient's relation to the clusters for key features. In this example we can see that for most features the patient is again most similar to cluster C. In order to obtain the presented values, we initially calculated the weighted average of the features' values for each cluster. Then, we measured the distance between the patient's feature values to each cluster. We then normalized the values to a scale between 0 and 1 and inverted them.*

# 4. Discussion

4.1 Summary

In this article, we demonstrated a method for deriving treatment-relevant patient subgroups defined in a manner that can be easily interpreted by clinicians. Using the DPNN, we generated a model which performed stably across many runs and which has the potential to increase remission rates (if used to select treatments for patients) by 6.5% at the population level (a 13.5% relative improvement), despite the fact that many of these treatments are roughly equivalently effective when considered at group level. Overall model performance was similar to both our previous use of the DPNN, as well as other papers predicting remission with antidepressants (Kleinerman et al., 2021; Meltretter et al., 2020a; Meltretter et al., 2020b; Kessler et al., 2022; Chekroud et al., 2016). In addition, real remission rates were higher when patients were assigned treatments with the highest output probabilities for them, suggesting that model estimates of treatment success probability do reflect improved chances of treatment success. In previous work we have argued that these results, especially those related to improving the population remission rate, are clinically meaningful (Mehltretter et al., 2020a). We will therefore focus the discussion on the three novel derived patient subgroups, which are differentiated in terms of their symptoms in a manner that appears to be clinically interpretable and as such may be useful to clinicians engaging with AI-enabled decision support systems.

4.2 Novel clusters

When examining the clusters, it is clear that they differ in terms of both overall remission rates and remission rates of treatments (Table 4). This finding implies that the DPNN is a potentially useful approach for the identification of novel and clinically meaningful patient subgroups. These patient clusters can be described narratively, providing a more intuitive approach to their interpretation. Cluster A is a group of younger patients with significant symptoms, increased fatigue, and reduced remission rates; their most effective observed treatments include both monotherapy and combination treatments. We call this the "Moderately severe, young and fatigued" group. Cluster B patients are more likely to be female and older, have less severe symptoms than other clusters and tend to respond better to a larger range of treatments. We call this the "Moderate, majority female" group. Finally, Cluster C patients are older, have reduced remission rates (similar to cluster A), more agitated depression, more sexual dysfunction, and somewhat increased suicidality. These patients may have an improved response with venlafaxine, and putatively less apparent tolerability for medications or combinations with more significant side effect profiles. We call this the "Severe, Agitated and Suicidal" group.

With respect to the clinical meaning behind these clusters, we can make some observations. Cluster C's increased agitation may explain why the treatment combination with bupropion is lower down in its treatment list; bupropion has a tendency to be less effective for anxious symptoms than SSRIs in some studies and, in rare cases, can worsen agitation and anxiety (though overall it remains an effective treatment for anxious symptoms) (Metaragno, 2021;Patel et al., 2016; Schatzberg & DeBattista, 1999). Cluster A also has two combination treatments in its top three treatments and the top treatment in cluster B is a combination, suggesting that patients in cluster A and B are more likely to be able to tolerate or benefit from combination treatment. In contrast, none of cluster C's top three treatments are combinations, suggesting patients in this cluster are less able to tolerate and benefit from treatment combinations. Cluster C also tends to have more suicidality which is in keeping with both the lower remission rate (Jeffrey et al., 2021) compared to cluster B and the increase in agitation, which is a risk factor for suicidality (Popovic et al., 2015). It is also interesting to note that venlafaxine was the most effective drug in cluster C; this finding is in line with a recent cluster analysis of venlafaxine effectiveness which demonstrated that patients with higher depression severity as well as scores on sexual dysfunction, agitation, and anxiety derived more benefit from venlafaxine (Kato et al., 2020). See the supplementary discussion for further discussion.

We note as well that cluster C was differentiated from the other clusters by its greater diversity in terms of race and ethnicity. We advise caution when interpreting these findings as our dataset was lacking in key data for social determinants of health, such as income and education. In previous analyses of subsets of this data, where these variables were available, we did not find race to be a useful predictor of outcome (Mehltretter et al., 2020b) and there is a lack consistent evidence of a direct relationship between race and outcome (Perlman et al., 2019). As such in more complete datasets we may expect this relationship to change.

4.3 Future Work

We believe it prudent to note that this model is not yet ready for clinical implementation; further validation in larger datasets and in subgroups is warranted. In addition, it is not yet clear if the subgroups demonstrated here are stable in time, or if they change at different phases of treatment, e.g. after treatment failure. Future work could address this. In addition, implementation science work is needed to better assess how these models would operate within the complex clinical environment; we have begun studying this in other work (Benrimoh et al., 2021; Popescu et al., 2021; Tanguay-Sela et al., 2022; Kleinerman et al., 2022). Our focus in this paper was generating a model that may eventually prove to be clinically useful, and as such we did not include placebo as a treatment option; future work may choose to include placebo in order to study more basic questions about placebo vs. treatment response. In addition, while in this work data was only available to compare pharmacological treatment options, future work may be able to extend this method to other treatment modalities. Future work with more appropriate datasets may also better investigate the underpinnings of the increased suicidality seen in cluster C. We also note that this work is focused on improving interpretability of machine learning models; it does not supplant, but is rather aimed to complement, existing methods of understanding relationships within data, such as moderation and mediation analyses. While models such as these may generate probabilities and classifications using learned statistical associations, they should not be construed as having defined causal relationships without more definitive experimental research and replication. Indeed, because this analysis is dependent on sequential retrospective analysis of data, these results should be viewed as potentially useful associations rather than definitive causal relationships. Most importantly, future work will serve to determine whether in clinical practice the use of this model and its accompanying visualizations will help improve interpretability for clinicians and enhance shared decision making by helping clinicians and patients both better understand model results. In addition, clinical studies utilizing model outputs will be necessary in order to determine if they can help improve patient outcomes, as the results of clinical studies may not necessarily line up with results from previous trials or current clinical best practices (Lenze et al., 2023).

### 4.4. Limitations

There are a number of limitations in this paper. One is a limited number of features, something which will need to be resolved by integrating larger datasets with more consistent feature sets in the future. The main consequence of this limitation is that the patient subgroups derived here cannot be considered 'final', and would need to be replicated in larger datasets with more available features prior to clinical use. Another limitation is the lack of key socio-demographic features, such as education and income, and of features such as medical comorbidities and frailty in the elderly. Another limitation is the lack of genetic or other biological data. Its absence does allow such a model to be more easily incorporated into clinical practice (Popescu et al., 2021), but it does reduce our ability to test hypotheses about the neurobiological underpinnings of observed inter-cluster differences. One other limitation is the lack of out-of-sample validation for all the medications in the dataset; see supplementary section 6 for further details. In addition, it should be noted that some of the datasets in this study were single arms of treatment studies, with all patients in these studies receiving the same treatment. While this reduces the risk of confounds such as those in naturalistic observational datasets, and the purpose of the study was to generate individual treatment forecasts rather than to generate group comparisons, careful comparison of these results with results from models trained on only randomized studies may be warranted in the future, as inferences from those datasets may be stronger (Furukawa et al., 2020; Furukawa et al., 2018). In addition, replication on datasets with more treatments within a single dataset will help to assess risk of confounding.

4.4 Conclusion

In conclusion, the results provided here are a first demonstration of the ability of the DPNN architecture to derive clinically meaningful patient subgroups based on treatment outcome, rather than purely on symptomatology. While the subgroups presented here cannot be considered as confirmed findings, they do already suggest several interesting avenues for novel research and demonstrate that patient subgroups generated by machine learning models may contribute to our ability to understand model outputs about treatment outcome both in research and clinical settings. In addition, improved interpretability as a result of this approach may help improve communication and shared decision making between clinicians and patients when using these models.

Data availability: The data used for this project was provided from the NIMH as well as by Dr. Furukawa and the University of Kyoto, as well as the IRL-GREY investigators and the University of Pittsburgh, for which we are very grateful. Data and/or research tools used in the preparation of this manuscript were obtained from the National Institute of Mental Health (NIMH) Data Archive (NDA). NDA is a collaborative informatics system created by the National Institutes of Health to provide a national resource to support and accelerate research in mental health. Dataset identifier(s): Sequenced Treatment Alternatives to Relieve Depression (STAR*D) #2148,
Combining Medications to Enhance Depression Outcomes (CO-MED) #2158,
Research Evaluating the Value of Augmenting Medication with Psychotherapy (REVAMP) #2153, Establishing Moderators/Biosignatures of Antidepressant Response - Clinical Care (EMBARC) MDD Treatment and Controls #2199. This manuscript reflects the views of the authors and may not reflect the opinions or views of the NIH or of the Submitters submitting original data to NDA. The authors do not have permission to share the data used in this project as it does not belong to them; those interested in using the data should contact the NIMH, the University of Pittsburgh, and the University of Kyoto.

Funding: Funding provided by ERA-PERMED supporting the IMADAPT project; the NRC via IRAP; and Aifred Health

Conflict of interest statement: DB, CA, JM, RF, KP, SI, CP, GG, SQ, AA were all employees, officers, shareholders, or contractors of Aifred Health when this work was done. JK is a member of an advisory board to Aifred Health and has received stock options. AK, AK and AR have all collaborated with Aifred Health in the past and have received honoraria. CR receives an honorarium from the American Association for Geriatric Psychiatry as editor of the American Journal of Geriatric Psychiatry; royalty income for intellectual property from the University of Pittsburgh (Pittsburgh Sleep Quality Index), from Oxford University Press, and from Up-to-Date; and consulting income from the University of Maryland, Weill Cornell College of Medicine, Washington University of Saint Louis, and the University of South Florida. TAF reports personal fees from Boehringer-Ingelheim, DT Axis, Kyoto University Original, Shionogi and SONY, and a grant from Shionogi, outside the submitted work; In addition, TAF has patents 2020-548587 and 2022-082495 pending, and intellectual properties for Kokoro-app licensed to Mitsubishi-Tanabe. EL reports grant support from the COVID early treatment fund, Fast grants, Janssen; consulting for Merck, IngenioRx, Prodeo, Pritikin ICR, Boehringer Ingelheim; and a patent pending for Sigma 1 agonists for COVID-19. Within the past five years, JK has received compensation for development and presentation of a (disease-state, not product-focused) webinar for Otsuka. He has served as scientific advisor (paid) to NightWare and Biogen and (with potential for equity) to AifredHealth. He receives compensation from Journal of Clinical Psychiatry and American Journal of Geriatric Psychiatry for editorial board service.

# Supplementary Material

1. The Dataset
    1.1 The clinical trial datasets
    1.2 Features in the Merged Dataset
    1.3 PCA
    1.4 Explanatory notes for SUN☺D
2. Additional Analysis of Clusters
    2.1 Features' Distribution Across Clusters
    2.2. Statistical Differences
    2.3. Further cluster descriptions
3. Additional Results
    3.1 Analysis of Results with the Sertraline Arm from SUN☺D
    3.2 Interpreting the Clusters with Decision Trees
    3.3 Additional visualizations
4. DPNN-Further Description
    4.1 Further description of cluster derivation
5. Supplementary discussion
6. Out-of-sample validation

# 1. The Dataset

1.1. The clinical trial datasets

STAR*D (N= 4041): The largest pragmatic trial of depression treatment. We used data from the first arm, where all patients received citalopram. Allowed patients with chronicity or treatment resistance.

CO-MED (N= 665): A study in which patients were randomized to three combination antidepressant treatment arms: escitalopram and placebo, bupropion and escitalopram, or mirtazapine and venlafaxine.

REVAMP(N= 808): A study comparing a medication with two different psychotherapies; we analyzed the medication only group (patients on escitalopram, bupropion, venlafaxine or mirtazapine). Allowed patients with chronicity or treatment resistance.

EMBARC(N= 296): A study focused on finding biomarkers of depression treatment response. We used the sertraline and bupropion groups.

IRL-GREY (N=468): A multiple stage study. We used the data from the first stage of the study, where all patients received venlafaxine monotherapy. Allowed patients with chronicity or treatment resistance.

SUN☺D (N= 2011): This study included two steps. First, patients received sertraline, and after 3 weeks, non-responsive patients received sertraline, mirtazapine, or add-on mirtazapine in a randomized manner. We used the mirtazapine and mirtazapine augmentation arms in the primary analysis, but also have a supplementary analysis where we include the sertraline arm (initially excluded because we know these patients had a poor initial response to the drug they were on).

| Clinical Trial | Number of patients |
| --- | --- |
| STAR*D | 2477 |
| SUN☺D | 1093 |
| REVAMP | 797 |
| CO-MED | 648 |
| IRL-GREY | 372 |
| EMBARC | 114 |

Supplementary Table 1a. The number of patients from each clinical trial that were included in our dataset

| Treatment | Number of patients |
|---|---|
| Citalopram | 2477 |
| Sertraline | 726 |
| Mirtazapine | 559 |
| Mirtazapine + sertraline | 536 |
| Venlafaxine | 402 |
| Escitalopram | 311 |
| Mirtazapine + venlafaxine | 214 |
| Bupropion + escitalopram | 213 |
| Total | 5438 |

Supplementary Table 1b. Number of patients per treatment in our dataset

1b. Features in the merged dataset

The first step in merging datasets involved creating the taxonomy, an iterative process of generating a tree diagram based on personal (i.e., sociodemographic) characteristics and clinical dimensions. Given the differences across the same questionnaires (e.g., different wording of questions, differing versions, etc…), a standard version for each questionnaire was created. We also created a taxonomy "tree" for the purpose of organizing standard questions into semantic categories. This structure was originally inspired by the IMAS classification system (cite) but was expanded greatly to encompass many dimensions of affect, lifestyle, psychiatric history, functional impairment, physical symptoms, etc.. Each standard question was assigned a taxonomical term, which was then matched to raw study variables in our database. Based on this matched standardized data, categories were generated with questions organized by taxonomy branches. We refer to these new categories as transformed questions. These transformed questions, which contained either categorical or binary response values, were based on semantically analogous questions. In order to maximize the amount of relevant information we could include in each of the transformed questions, we would often "binarize" questions that had categorical response values. We created thresholds for each question to map categorical responses into binary values. Manual decisions for cutoffs for binary questions made based on group consensus and were verified by assessing similarity between the binary response value distributions from originally binary questions (e.g., no mapping needed, only had 2 response values) and the mapped distributions from originally categorical response value questions. For transformed questions that included questions with only categorical response values, equipercentile equating was used as a standardization. To account for missing data, we used MICE as an imputation method and then averaged the imputed datasets. Note that in accordance with best practices we chose not to impute data that was systematically missing from certain datasets. Finally, more standardized scaling was conducted to produce a final set of transformed questions used as the input for model feature selection.

As mentioned in the article, after merging the datasets together, we were left with 19 common features. Supplementary Table 2 presents the features, divided into categories.

| Symptom features | <ul><li>Total baseline severity (QIDS total score or study equivalent) QIDS-SR-16</li><li>Suicidal ideation and planning</li><li>Guilt</li><li>Feelings of worthlessness</li><li>Psychomotor agitation</li><li>Genital symptoms (loss of libido, menstrual disturbances)</li><li>Anhedonia</li><li>Sadness</li><li>Fatigue/lassitude</li><li>Overall suicidal ideation</li><li>Guilt – binary</li><li>Anhedonia – binary</li><li>Negative thoughts (binary version 1)</li><li>Marked negative thoughts ( binary version 2)</li><li>Feelings of worthlessness – binary</li><li>Excessive guilt – binary</li></ul> |
|---|---|
| Demographic features | <ul><li>Age</li><li>Sex</li><li>Race/ethnicity</li></ul> |

Supplementary Table 2. The features in the dataset (after merging)

The merged dataset initially included 9 courses of treatment: citalopram, sertraline, mirtazapine, mirtazapine combined with sertraline, venlafaxine, escitalopram, mirtazapine combined with venlafaxine, bupropion combined with escitalopram, and bupropion monotherapy. During model development, we removed one of the original treatment courses, bupropion monotherapy, because of low patient numbers (n=63). Therefore, the dataset we finally used in the evaluation included 8 treatments.

1.3. The Principal Component Analysis (PCA) for Preliminary Validation of Dataset Merging

As preliminary validation, we first investigated the distribution of the patients' samples after transforming the feature values and merging them all into a single dataset. Our objective was to verify whether studies were easily separable by the included features (if patients post-merging cannot be easily grouped into their respective studies, this suggests the merging created a uniform dataset). This was important in order to validate the quality of the feature transformation and to prevent the model from learning prototypes that represent individual studies (and that are therefore not informative).

We performed a principal component analysis (PCA) in order to determine if the data post-transformation showed clear structure related to study. Of note, this version of the PCA was done with 18 of the 19 available patient level features as well as the outcome (remission) for a total of 19 variables included in the PCA; we did not include the race/ethnicity variable given that the SUND study was essentially only comprised of Asian patients and as such we expected it to be differentiated on this basis; as can be seen in the decision tree results below and the feature distributions discussed in the main paper, however, it does not appear that Asian ethnicity generated its own cluster in terms of the results. For completeness, we also include a version of the PCA with race and ethnicity below.

We used the first two components in order to visualize the distribution of the patients. As can be seen in Supplementary Figure 2, there is clear overlap between patients from different studies when visualizing the first two components, indicating successful dataset merger. See below for more details about the PCA.

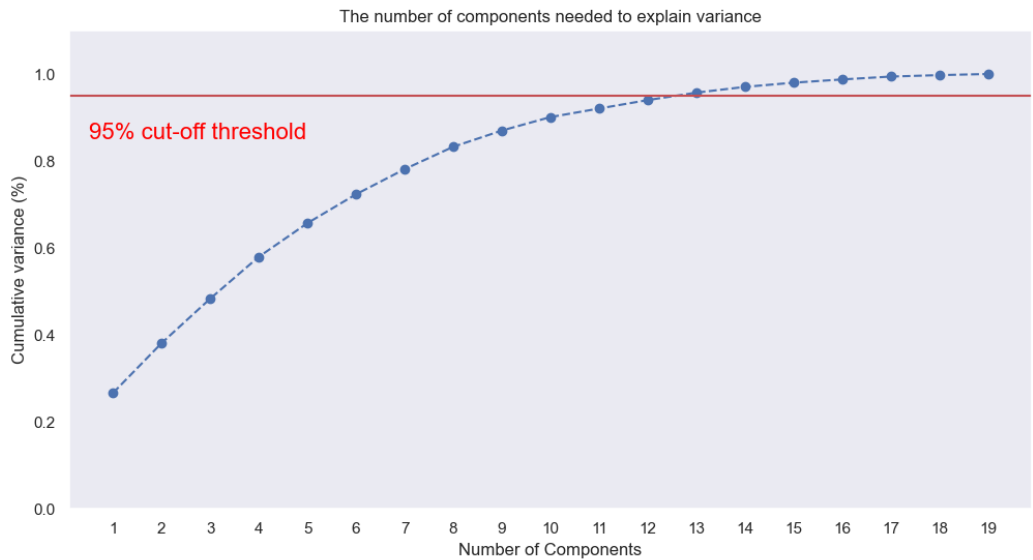

Supplementary Figure 1. The number of components needed to explain the variance of the 19 original features.

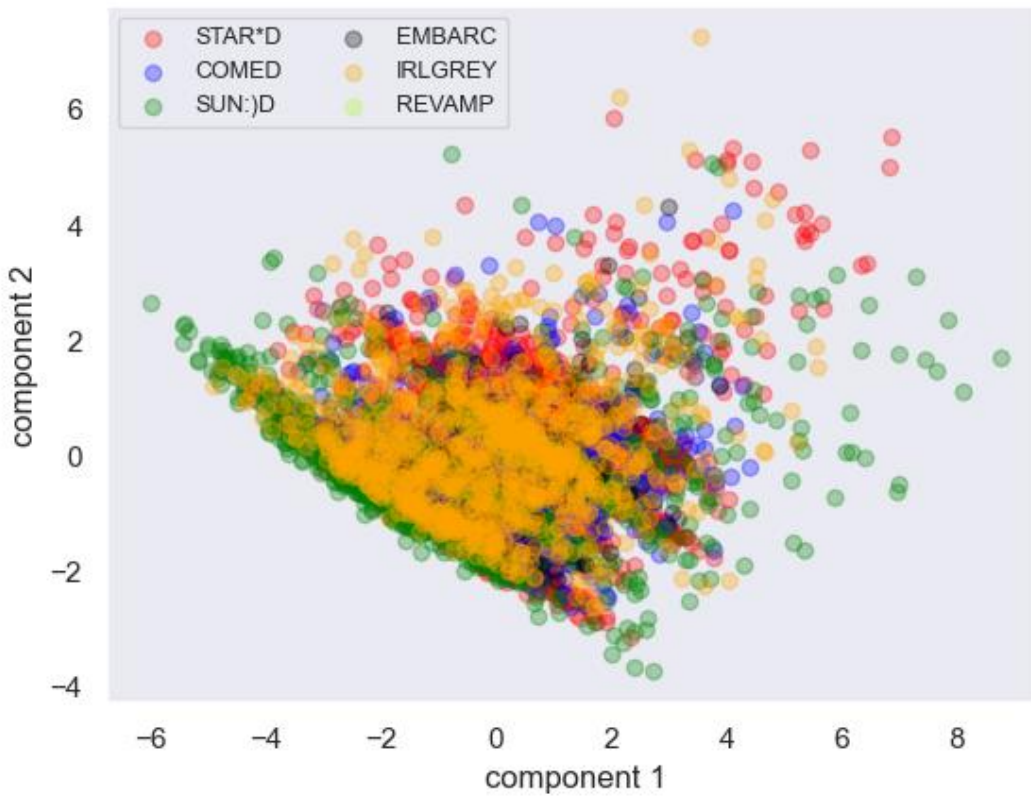

Supplementary Figure 2. PCA-based visualization of patient distribution, color-coded by study

In our PCA analysis, we focused on the two first components. The values of the components for the original features are presented in Supplementary Table 3.

| Original feature | Component 1 | Component 2 |
|---|---|---|
| Age | -0.062 | 0.082 |
| Total severity score | 0.446 | -0.119 |
| Suicidal ideation and planning | 0.381 | 0.573 |
| Guilt | 0.278 | -0.201 |
| Feelings of worthlessness | 0.277 | -0.104 |
| Psychomotor agitation | 0.197 | -0.186 |
| Genital symptoms (loss of libido, menstrual disturbances) | 0.154 | -0.196 |
| Anhedonia | 0.277 | -0.302 |
| Sadness | 0.328 | -0.13 |
| Fatigue | 0.276 | -0.292 |
| Overall suicidal ideation | 0.38 | 0.566 |
| Remission | -0.061 | 0.016 |
| Sex | 0.018 | -0.051 |
| Guilt binary | 0.087 | -0.059 |
| Anhedonia – binary | 0.041 | -0.043 |
| Negative thoughts – binary. version 2 | 0.075 | -0.025 |
| Negative thoughts – binary. version 1 | 0.039 | -0.013 |
| Feelings of worthlessness – binary | 0.063 | -0.029 |
| Excessive guilt – binary | 0.079 | -0.051 |

Supplementary Table 3. The values of the top two components in the PCA

We also investigated how many components were necessary to sufficiently explain the variance of the features. We obtained the following chart (Supplementary Figure 1). As can be seen, many components are needed to explain a significant amount of the variance, suggesting that the data is not easily separated into groups based on the included features.

We focused on the first 2 components in order to provide a visualization of the results.

PCA With Race/ethnicity included

Here for completeness we provide the PCA including race/ethnicity. While the graph of the first two components does show more separation between studies, there remains significant overlap, and the explained variance continues to depend on many components, suggesting that adding race/ethnicity does not significantly increase the explained variance of single components.

Supplementary Figure 1a. The number of components needed to explain the variance of the 19 original features, including race/ethnicity

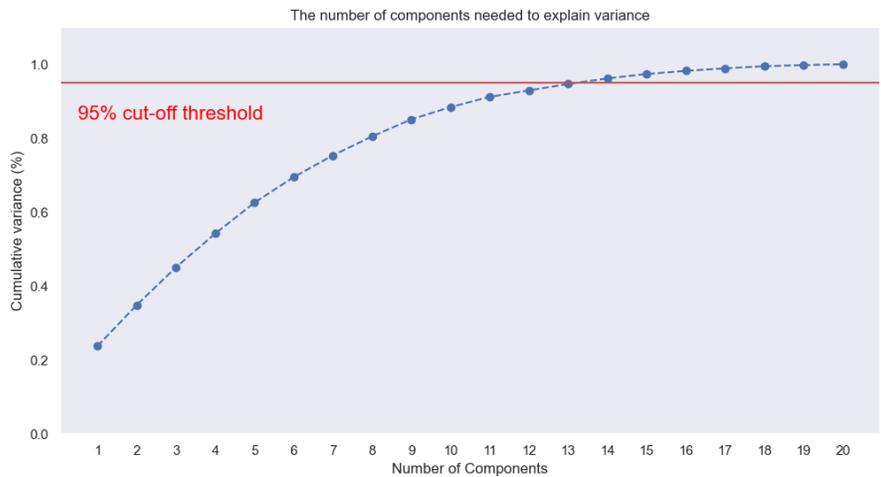

Supplementary Figure 2. PCA-based visualization of patient distribution, color-coded by study, including race/ethnicity

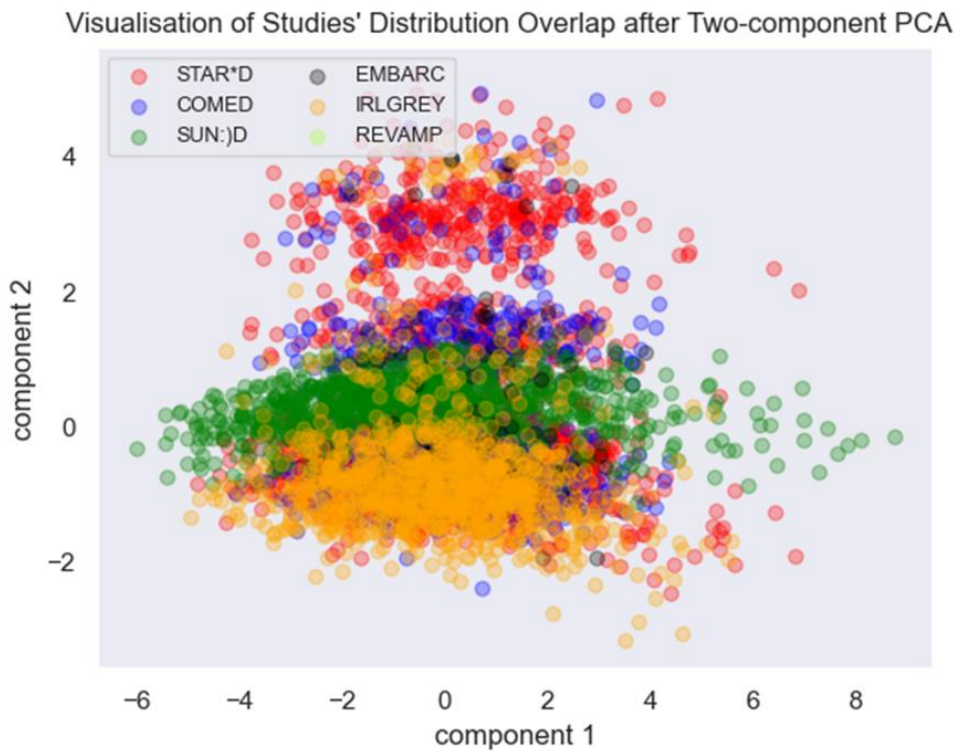

Supplementary Table 3a: The values of the top two components in the PCA including race/ethnicity

| Original feature | Component 1 | Component 2 |
|---|---|---|
| Age | -0.063 | -0.223 |
| Total severity score | 0.446 | 0.055 |
| Suicidal ideation and planning | 0.381 | -0.134 |
| Guilt | 0.278 | 0.019 |
| Feelings of worthlessness | 0.276 | -0.12 |
| Psychomotor agitation | 0.197 | 0.114 |
| Genital symptoms (loss of libido, menstrual disturbances) | 0.155 | 0.19 |
| Anhedonia | 0.277 | 0.022 |
| Sadness | 0.328 | 0.026 |
| Fatigue | 0.276 | 0.039 |
| Overall suicidal ideation | 0.379 | -0.149 |
| Remission | -0.061 | -0.041 |
| Sex | 0.019 | 0.024 |
| Race/ethnicity | 0.021 | 0.915 |
| Guilt binary | 0.087 | -0.014 |
| Anhedonia – binary | 0.04 | -0.002 |
| Negative thoughts – binary. version 2 | 0.075 | -0.012 |
| Negative thoughts – binary. version 1 | 0.039 | -0.013 |
| Feelings of worthlessness | 0.063 | 0.004 |
| – binary | | |
| Excessive guilt – binary | 0.079 | -0.005 |

1.4. Explanatory notes for SUN☺D

For each dataset we focused on the first treatment utilized; in SUN☺D, where treatments were changed relatively early (at 3 weeks) for those not initially responding to sertraline, we included patients who were randomized to mirtazapine or to mirtazapine-sertraline combination therapy. This may have reduced the apparent effectiveness of mirtazapine as the patients receiving them had failed a treatment within the current episode and is mitigated by two factors. The first is that other datasets, such as STAR*D and REVAMP included patients with chronic depression and previous treatment failures, and as such even if these treatment failures were not directly captured in the data, portions of the patient populations in these studies did have treatment resistance. The second is that the primary purpose of this paper is to demonstrate how clusters can be derived and interpreted; confirmatory analyses in larger samples with more coverage of the different medications would be required to make definitive statements about the clinical validity of the clusters derived in any case.

## 2. Additional Analysis of Clusters

2.1. Feature Distribution Across Clusters

For non-binary features:

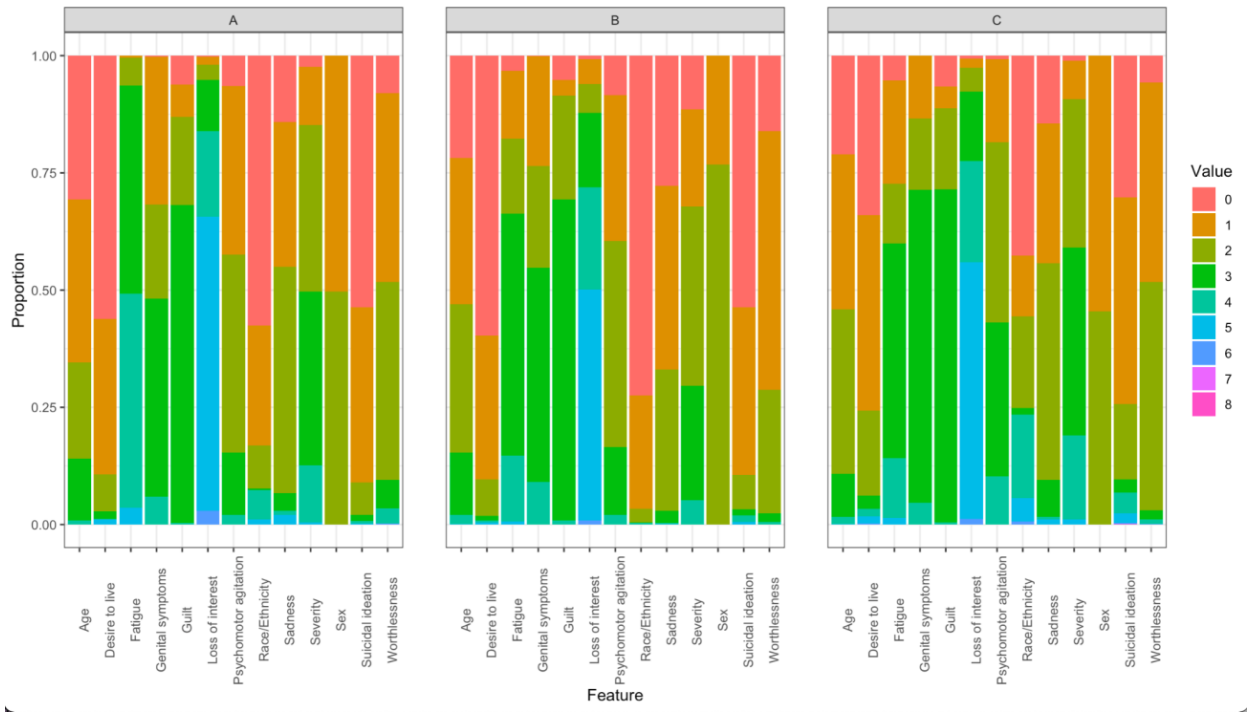

For binary features:

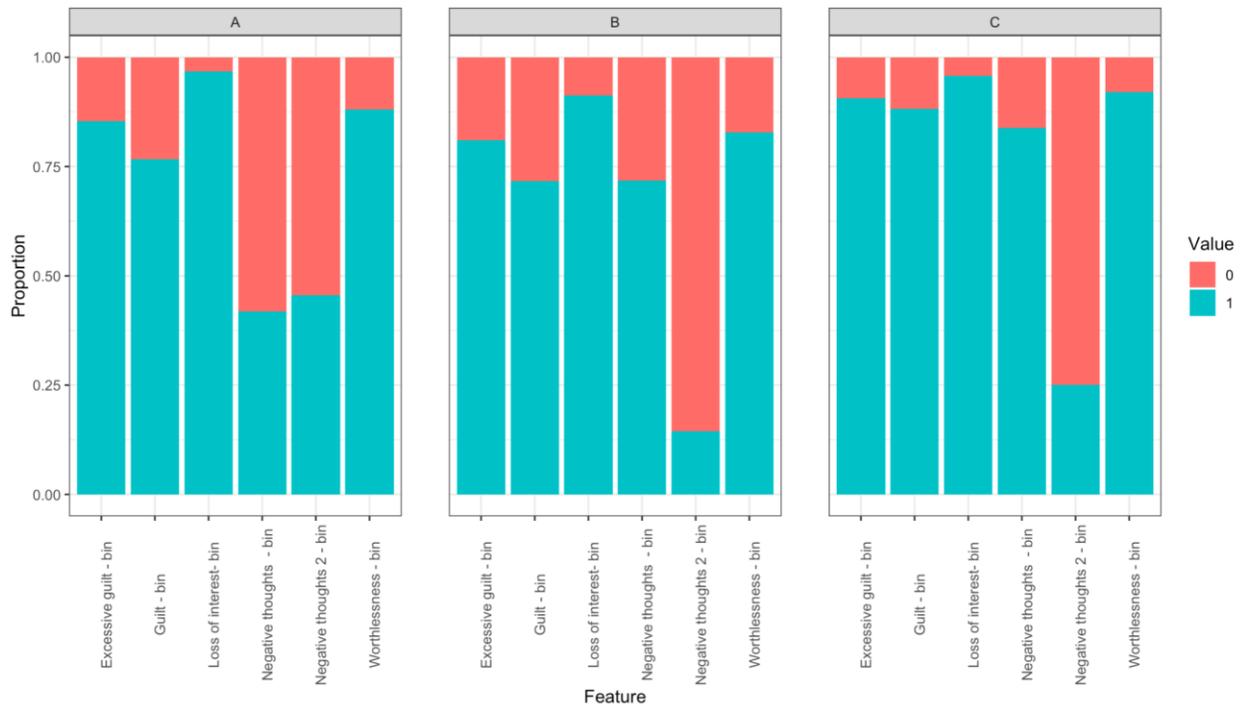

## 2.2. Cluster Statistical Analysis

In this subsection we present a statistical analysis of the differences between the clusters in all the values of all features.

First, in order to determine the appropriate statistical test, we examined if the feature values were. Using the Shapiro-Wilk, we found that many of the values were not normally distributed (Royston, 1992). Therefore, we used Kruskal–Wallis test (McKight and Najab, 2010), which does not assume normality, and Dunn's test (Dunn, 1964) was used for post-hoc pairwise comparison (McKight & Najab, 2010; Dunn, 1964).

The Kruskal–Wallis test showed that the clusters were significantly different ($p<0.01$) from each other for the values of all features except for guilt ($p = 0.13$). In the post-hoc Dunn's test we found that the differences between clusters A and C for the following features were *not* significant or only marginally significant: Sadness ($p= 0.06$), Feelings of worthlessness ($p=0.06$), Guilt($p=0.18$). In all the other features the differences were significant ($p<0.01$). As this was an exploratory analysis we did not correct for multiple comparisons.

Relevant references:

## 2.3 Further cluster descriptions:

| Treatment \ Cluster | A | B | C |
|---|---|---|---|
| Citalopram | 582 | 1180 | 715 |
| Escitalopram | 104 | 132 | 75 |
| Escitalopram +bupropion | 80 | 71 | 62 |
| Mirtazapine | 222 | 256 | 81 |
| Mirtazapine+ sertraline | 175 | 304 | 57 |
| Sertraline | 306 | 300 | 120 |
| Venlafaxine | 199 | 140 | 63 |
| Venlafaxine+ mirtazapine | 74 | 76 | 64 |
| Total size | 1742 | 2459 | 1237 |

Supplementary Table 4. Number of patients per cluster and treatment type

In most of the symptom features, cluster C is relatively similar to cluster A, though there are some differences between these and cluster B. One prominent exception to the similarity in symptoms between cluster C and A is in psychomotor agitation; here, the severity of the symptom in cluster C is higher than in cluster A (and also cluster B). Other observations can be made. Cluster B has more females, compared to clusters A and C which have more even sex distributions. The severity score in cluster C is overall similar to cluster A, and both are more severe than cluster B. Cluster C tends to also demonstrate more diverse ethnicities than the other clusters. Despite their similarities, cluster A and cluster C have different reactions to treatments. Cluster A was most likely to achieve remission with citalopram (45% remission rate), while cluster C benefited most from venlafaxine(49%). The highest remission rate in cluster B, of 62%, was found for escitalopram and bupropion. From a clinical perspective, it is interesting to observe the individual symptoms that differentiate the clusters. One such symptom is psychomotor agitation, which tends to be higher in cluster C than in A or B; as we will discuss below, this is perhaps why this cluster has the combination including bupropion lower down in the list of treatments than the other clusters. Cluster C has more severe somatic genital symptoms as well (e.g. loss of libido), and patients in Cluster A tend to have somewhat higher severity of fatigue. Cluster C also seems to have slightly increased suicidality and somewhat less fatigue.

# **3. Additional Results**

## 3.1. Analysis of Results with the Sertraline Arm from SUN☺D

As mentioned above, the SUN☺D trail included three arms. One arm received (after three weeks of receiving sertraline) mirtazapine, one arm received sertraline with mirtazapine, and the third arm only received sertraline. In the primary analysis, we did not include the arm who only received sertraline. Supplementary Table 4 presents the analysis including the sertraline only arm.

| Number of prototypes | Area under curve (AUC) | Sensitivity | Specificity | Positive predictive value (ppv) | Negative predictive value (npv) | F1* | Remission Rate** |
|---|---|---|---|---|---|---|---|
| 3 | 0.642 | 0.52 | 0.676 | 0.502 | 0.674 | 0.501 | 0.38 |
| 4 | 0.644 | 0.485 | 0.692 | 0.507 | 0.673 | 0.496 | 0.417 |
| 5 | 0.647 | 0.479 | 0.668 | 0.433 | 0.641 | 0.483 | 0.399 |

Supplementary Table 5. The accuracy results of the DPNN model with the data of patients who received sertraline only in the SUN☺D trial

## 3.2 Interpreting the Clusters with Decision Trees

After deriving the clusters and analyzing the distribution of the features, we used a decision tree classifier to further interpret the clusters (Loh, 2011). The aim here was to determine which features differentiate between the clusters.

Although decision trees may not be completely accurate, they are easy to interpret and can provide additional insights into the clusters learned by our neural network. Below we present a four-level decision tree for classifying a patient into one of the three clusters discussed above (Figure 4). The decision tree is 71.6% accurate. However, deeper decision trees will naturally give higher accuracy (we observed a maximum of 80% without limiting depth). Here we only analyze the four-level tree for the sake of simplicity. The decision tree in Figure 5 measured the quality of the splits using the gini impurity measure (Suthaharan, 2016).

Interestingly, we can observe that the node in the root of the tree splits according to the symptom of fatigue. Additional features that are considered early in the classification by the tree are race/ethnicity, overall severity score, negative thoughts, feelings of worthlessness, age, and desire to live.

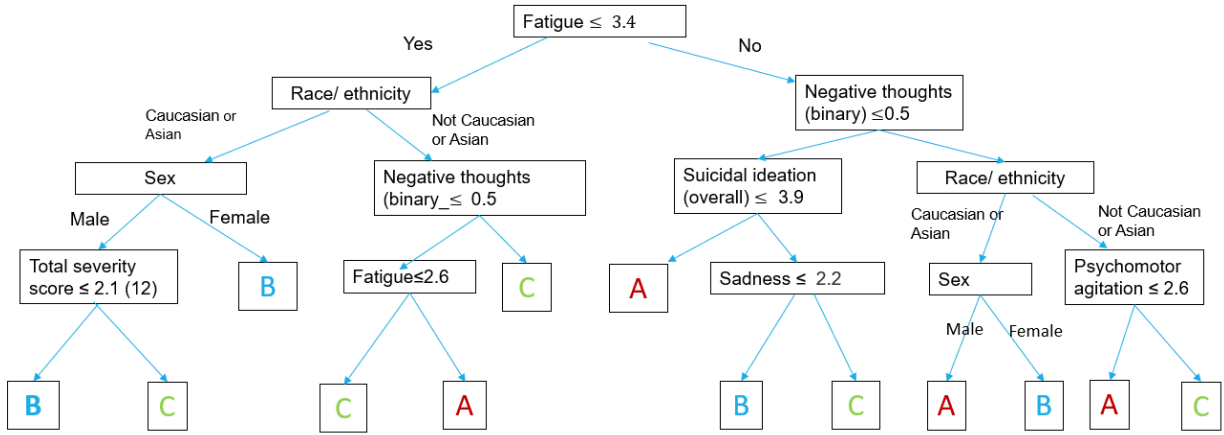

Figure 5. Decision trees trained to separate clusters

We used a series of 100 consecutive runs in order to investigate the stability of the clusters and specifically the decision trees. We measured how often the various features appeared within the top 3 levels of the tree. We found that some features appeared in the top of the trees frequently, while others rarely appeared. Supplementary table 6 presents the frequency of the 5 most frequent features in the decision trees:

| Feature | Frequency in Decision Trees |
|---|---|
| Race/ethnicity | 0.65 |
| Overall suicidal ideation | 0.54 |
| Total baseline severity | 0.42 |
| Guilt | 0.34 |
| Negative thoughts (binary) | 0.32 |

Table 5. Frequencies for the 5 most frequent features in the decision trees

As such, we can demonstrate that there is partial conservation of key discriminative features, meaning that decision tree models may be a helpful adjunct to interrogate the features that drive clustering, though this may be most helpful when considering results from several models rather than a single model, given the variability in which features are most commonly retained by the decision tree models.

*Cluster Interpretation with Decision Trees*

The decision trees provide another way to investigate the results of these subgroups. Firstly, it must be understood that these trees are imperfect, *post-hoc* models of subgroups are themselves derived from latent space representations. However, they may provide intuitive ways for clinicians to interpret investigations and results for a given patient. For example, a clinician could trace a patient's subgroup assignment using a decision tree as a means of contextualizing relevant aspects of the patient's symptomatology or demographics and their evidence-based role in that patient's treatment outcome. This may be a useful visualization technique for appreciating patient data besides simply telling clinicians to which group a given patient belongs. In addition, the decision trees confirm some of the information gleaned from examining the features we presented. For example, race, overall baseline severity, and a measure of suicidality (which is also an indicator of severity of pathology as explored in Corral et al., 2022) are three features which frequently appear in the top levels of the decision trees - in line with the finding that overall severity seems to distinguish clusters A and C from B and also that race/ethnicity distinguishes cluster C from A and B (potentially as a proxy for other social determinants of health which we unfortunately did not have access to in our dataset).

This differentiation of cluster C from A and B on the basis of diversity in terms of race and ethnicity should be interpreted with extreme caution: in previous work (Mehltretter et al., 2020) with a subset of this data, where both race as well as education and income data were available, we did not find race to be a predictive feature; in addition, there is no consistent evidence of a direct effect of race on treatment outcome (Perlman et al., 2019). As such, it is likely reasonable to assume that, in a more complete dataset with more detailed sociodemographic information, ethnicity would likely be shown to be acting as a proxy for other social determinants of health.

Some of the notions arising from these subgroups, such as the finding that bupropion may be less effective in cases of increased psychomotor agitation and that overall baseline severity reduces remission rates, have previously been observed in the literature as discussed. The replication of such findings as important aspects in the decision tree is reassuring as it provides some external validity to our results. In future work, we hope to present these novel patient subgroups, as defined by patient outcomes to treatment, in ways that help clinicians better interpret model predictions and utilize them to improve patient care.

3.3. Additional visualization supplementing main text figure 3:

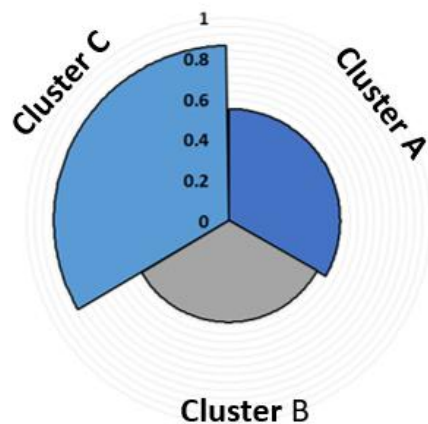

Figure 3b: Visualization of a patient's relation to the clusters (patient drawn from the dataset). The shaded area represents the similarity of the patient to the various prototypes, according to the Euclidean distance in the latent space calculated by the DPNN model. We normalized the values to a scale from 0 to 1 and inverted the transformed value. In this example, the patient is most similar to the prototype of cluster C.

# 4. DPNN Description

Similar to common machine learning models, the DPNN training loop aims to reduce a loss function. The loss function of the DPNN aggregates three components: 1) an accuracy loss function that aims to optimize the prediction of the outcome (in our case, the prediction of whether or not a patient will go into remission with a given treatment); 2) an autoencoder loss function, that ensures that the encoding produces meaningful latent features; and 3) prototype variance loss, that aims to increase the discordance between the prototypes with respect to their expected outcomes across the possible treatments. In our previous paper, Kleinerman et al. 2021, we demonstrated improved accuracy when using DPNN compared to a direct prediction of remission rates.

The prototype variance loss component enables the DPNN to potentially reveal prototypes that will be useful for treatment selection. This loss component aggregates two types of variance: the total variance of remission across prototypes and the total variance across treatments within the prototypes. The variance across prototypes is used to obtain prototypes that are different from each other, and the variance within prototypes is used to produce *meaningful* prototypes, with clear differences in the treatments effect (for each prototype).

4.1 Model Hyperparameter Tuning

Given that our goals were to optimize the accuracy and increase the interpretability of our model, it was necessary to tune the hyperparameters (including the number of prototypes). For this purpose, we followed an iterative process: we first used a grid search for parameter optimization in order to increase accuracy and then proceeded to analyze the prototypes together with content experts (psychiatrists) in order to estimate the interpretability and clinical value of the prototypes.

The number of prototypes was selected based on pragmatic considerations. There are currently 3 major commonly encountered subtypes of depression, based on symptomatology: melancholic depression, atypical depression, and depression with anxious distress (though we note these subtypes have been shown to be of limited value when predicting antidepressant response (Arnow et al., 2015)). As such, any number of prototypes less than three would be unlikely to provide much in the way of novel clinical information, and would most likely result in a redundant simple split between patients with good and poor outcomes. On the other hand, more than 5 prototypes, even if valid, would likely be too practically cumbersome to be clinically useful or to increase interpretability for the average clinician (this upper limit was based on the clinical experience and content expertise of the authors). As such, we focused on generating results using the number of prototypes expected set at 3, 4 and 5 prototypes.

References:

Arnow, B. A., Blasey, C., Williams, L. M., Palmer, D. M., Rekshan, W., Schatzberg, A. F., Etkin, A., Kulkarni, J., Luther, J. F., & Rush, A. J. (2015). Depression Subtypes in Predicting Antidepressant Response: A Report From the iSPOT-D Trial. American Journal of Psychiatry, 172(8), 743–750. https://doi.org/10.1176/appi.ajp.2015.14020181

4.2 Further description of cluster derivation

In addition to the remission prediction, the model outputted the final prototypes that were tuned during the training process. The prototypes consisted of features encoded in the latent space and therefore could not be interpreted directly. The full procedure for generating prototypes and then deriving clusters from the prototypes can be visualized in Figure 1. It is important to note that the k-fold testing process described above was intended to demonstrate adequate accuracy metrics. However, in the exploratory analysis of the prototypes, our interest was to determine their interpretability and potential clinical utility, and, as such, we use the DPNN on the whole dataset to maximize available data. This is done because the prototypes themselves are the focus of this analysis rather than accuracy (which requires a test set). We first trained and tested the DPNN model on the whole dataset. Then, we obtained the latent space representation of the prototypes that were tuned by the model during training. We then found, for each prototype, the group of patients that were closest to that prototype in the latent space. For this purpose, we first encoded all the patients with the DPNN's encoder to the latent space representation, and then calculated the Euclidean distance between each encoded patient to all prototypes. Then we associated the patient with the cluster of the closest prototypes. Thus, we obtained a number of clusters equal to the number of prototypes.

In order to further interpret the clusters and the differences between them in terms of effectiveness of medication, we obtained the size of the clusters and the distribution of medications actually received by the patients in the clusters. For each cluster, we calculated the average real remission rate for each medication which, in turn, generated a ranking of medications within each cluster.

We then utilized two methods to illustrate the clusters' features and the differences between them. First, we visualized the distribution of all feature values in all clusters using a histogram, which allows for comparisons between clusters. Second, we used a decision tree classifier trained on the clustered data in order to predict cluster membership (see supplementary section 3).

# 5. Supplementary Discussion

One important takeaway from this work is the practicability of aggregating data from different clinical trial datasets in order to generate a coherent model with accuracy comparable to models trained on more closely related datasets (Mehltretter et al., 2020). This is particularly noteworthy because the sample sizes of clinical trials are generally small from the perspective of machine learning, especially when compared to much larger datasets such as electronic medical records (EMRs). However, as we have discussed in other papers (Benrimoh et al., 2018), clinical trial datasets have certain advantages over EMRs, including clear outcome measures — which are imperative for the generation of targets for machine learning but are often lacking in EMRs. Moreover, while both EMR and clinical trial datasets can be biased (Braunholtz et al., 2001; Rubinger et al., 2022), clinical trials have at least part of their sampling bias noted explicitly, in the form of inclusion and exclusion criteria, and their use of randomization helps to remove bias related to treatment selection by clinicians (Sauer et al., 2022). While we discuss the methods we use for dataset integration in more detail in another paper (Mehltretter et al., 2020), it is important to note that this method seems to generate training data which can appear, after integration, as if it came from the same source dataset (see Figure 2).

The major drawback to the integration of disparate clinical trial datasets is the inevitable loss of features due to the lack of consistency between datasets and the limited overlap with respect to the measures contained therein. In our work, this scarcity of similar measures across trials resulted in a mere 19 features (from 72 features that were initially in our datasets separately) on which to train our model, so few that employing feature selection techniques did not seem to improve model performance. This phenomenon is distinct from our previous modeling work wherein larger feature sets were available and feature selection was foundational for improving model quality (Desai et al., 2021; Meltretter et al., 2020; Meltretter et al., 2020). Nevertheless, we made the decision to proceed with this diminished feature set primarily because it allowed us to include mirtazapine, which was not available in other datasets. The inclusion of mirtazapine is a relevant consideration when generating treatment subgroups that are based on outcome and differences in most effective observed treatments because it is an important first-line treatment with a different mechanism of action from other antidepressants (Benjamin & Doraiswamy, 2011; Jilani et al., 2022).

This limitation in the number of features likely limited the possible differentiation between subgroups.

*Model Performance and Cluster Observations*

One finding of note is the similarity between the models containing different numbers of prototypes. As discussed, this may be due to the restricted feature space: a larger feature space may lead to more significant differences between numbers of prototypes. Given this lack of clear differentiation between models containing different numbers of prototypes, we chose to focus on the three-prototype model.

With respect to symptomatology, one aspect that may help us understand the differing benefit of various treatments is the increased severity of psychomotor agitation in cluster C. This finding may explain why the treatment combination with bupropion is lower down in the treatment list for cluster C compared to the other clusters. Bupropion has a tendency to be less effective for anxious symptoms than SSRIs in some studies and, in rare cases, can worsen agitation and anxiety (though overall it remains an effective treatment for anxious symptoms) (Metaragno, 2021;Patel et al., 2016; Schatzberg & DeBattista, 1999). Cluster C also tends to have less fatigue than clusters B and C, providing another reason why bupropion (which tends to be helpful for fatigue as found by Kennedy et al., 2016) is less highly ranked. Cluster A also has two combination treatments in its top three treatments and the top treatment in cluster B is a combination, suggesting that patients in cluster A and B are more likely to be able to tolerate or benefit from combination treatment. In contrast, none of cluster C's top three treatments are combinations, suggesting patients in this cluster are less able to tolerate and benefit from treatment combinations. Cluster C also tends to have more suicidality which is in keeping with both the lower remission rate (Jeffrey et al., 2021) compared to cluster B and the increase in agitation, which is a risk factor for suicidality (Popovic et al., 2015).

It is also interesting to note that venlafaxine was the most effective drug in cluster C; this finding is in line with a recent cluster analysis of venlafaxine effectiveness which demonstrated that patients with higher depression severity as well as scores on sexual dysfunction, agitation, and anxiety derived more benefit from venlafaxine (Kato et al., 2020). Indeed, it is striking that the difference between the best treatment (venlafaxine) and the second best treatment (escitalopram) for cluster C is 10% - far larger than the equivalent difference in the other two clusters (2% in cluster A, 3% in cluster B), suggesting that the advantage for venlafaxine in this cluster is clinically significant.

If these clusters were confirmed in future analyses, clinicians could use the position of each patient relative to each cluster to better understand how they might respond to different treatments, and use subgroup membership to better understand the nature of the patient's situation and as an intuitive aid to improve their understanding of model predictions (Fig. 4). For example, if a patient is closest to cluster C, this might prompt a clinician to be particularly careful about treatment choice and to consider the role of psychomotor agitation in their overall symptomatic picture.

*Idiographic vs. Nomothetic approaches and the DPNN*

In our first use of the DPNN, we demonstrated that it performed better, in terms of improving population remission rates, than other machine learning models, including a deep learning model that did not learn patient prototypes (Kleinerman et al., 2021) - suggesting that there is potentially relevant information in patient prototypes which may be useful in predicting treatment outcomes. This is reminiscent of the dialectic between idiographic and nomothetic approaches to understanding psychopathology (Fisher et al., 2021; Reynolds, 1982; Reynolds, 1982; Reynolds, 1989; Wright & Woods, 2020). The idiographic position holds that only an understanding of the complex phenomenon that is the individual can provide us with the insight required to understand their clinical reality and the treatments most likely to be effective for them. While this approach would provide the most personalized interpretation of a patient's unique situation, it is impractical to accomplish on a large scale due to our limited understanding of the mechanisms underlying psychopathology and how these interact, which make it difficult to truly grasp a given patient's clinical situation based on first principles. Instead, we often rely on a statistical understanding based on the study of groups or subgroups of patients. This nomothetic approach is far more scalable and easier to use for the discovery of trends which may have relevance for individual patients. However, by looking at aggregated behavior, the nomothetic method necessarily sacrifices some resolution (i.e. the complex uniqueness of any given patient). The DPNN is an attempt to walk the line between these two approaches. On the one hand, it is idiographic in nature given that the model learns to make predictions for an individual patient, based on data from that individual patient. On the other hand, as the model is trained; it learns a potential structure within the patient population, as captured by the patient prototypes, which provide useful context when predicting results for individual patients. While the DPNN certainly does not aim to resolve the tension between idiographic and nomothetic approaches, it is intended as a clinically meaningful step in this direction.

*A note on calibration:*

As the main question was about the ability of the model to generate meaningful clusters, we did not explicitly evaluate calibration aside from the values provided for sensitivity, specificity, NPV and PPV, and F1. In future work, when designing clinic-ready models, we plan to employ further calibration metrics such as the MCE and ECE. However, as real remission rates were higher when patients were assigned treatments predicted as being most likely to be effective for them by the model, it seems as though model estimates of treatment success probability do reflect truly improved chances of treatment success. We note that while any model is capable of mis-classification, the use of the RRI metric for evaluation helps to allay this as it focuses on the remission rate of patients who received the drug predicted by the model to have the highest probability for success based on estimated the probability of remission, rather than the outputted classification; as such, model classification predictions are not used when calculating the RRI, but rather the probabilities predicted for individual treatments are used.

*A note on model deployment*:

This model, once further validated in a larger sample size, would likely be deployed in a similar way to our previous models (See Benrimoh et al., 2021, Tanguay-Sela et al., 2022 and Popescu et al., 2021) via a web-app which would intake patient questionnaire data, query a frozen version of the trained model, and then output results for clinicians in the form of a list of probabilities for all predicted treatments. As these kinds of models are classified by both Health Canada and, increasingly, the US FDA as medical devices, careful validation work must be undertaken prior clinical implementation.

## 6. Supplementary out-of-sample validation

With respect to validation, one limitation of the results presented is the lack of out-of-sample validation for our main results as a test of generalizability. This was due to the fact that there was poor overlap between medications in the datasets, meaning that removing any dataset as a whole from the training data would generally mean that the trained model would be tested on a medication it had not trained on, or trained on very little. The exception to this was sertraline, which existed in two datasets with sufficient data to attempt out-of-sample validation (REVAMP and EMBARC). We present an analysis of model performance where EMBARC serves as a held-out test set. The remission rates for sertraline were different between these studies, likely due to patient characteristics (e.g. chronic depression in REVAMP), and the EMBARC dataset was relatively small (n=114), and as such formed a small test set; for both of these reasons, we expected performance to suffer compared to the main model. Results indeed demonstrate reduced performance, but still demonstrate reasonable generalization (AUC close to 0.6, similar PPV, NPV and F1 values compared to the main model) given the limitations discussed. With respect to remission rate, the remission rate in the EMBARC dataset with sertraline was high (0.53) compared to sertraline in the REVAMP dataset (0.31) and the entire dataset (0.42). As such the low remission rate predicted here is likely a result of the model learning lower remission rates for sertraline from REVAMP. This again demonstrates the limitations of this held-out test set.

Table 6: Results for testing on the held-out EMBARC test set:

| AUC | Sensitivity | Specificity | ppv | npv | F1-score | Rem. Rate |
|---|---|---|---|---|---|---|
| 0.58 | 0.625 | 0.5429 | 0.385 | 0.76 | 0.4762 | 0.403 |

Table 7: Results for the cross-validated 3 prototype model (from Table 2 in the main text; for reference)

| Number of prototypes | Area under curve (AUC) | Sensitivity | Specificity | Positive predictive value (ppv) | Negative predictive value (npv) | F1* | Remission Rate |
|---|---|---|---|---|---|---|---|
| 3 | 0.666 (0.07) | 0.441 (0.3) | 0.731 (0.18) | 0.385 (0.24) | 0.678 (0.06) | 0.4 (0.26) | 0.48 (0.05) |